\documentclass[journal,twoside,web]{ieeecolor}
\usepackage{jsen}
\usepackage{cite}
\usepackage{amsmath,amssymb,amsfonts}
\usepackage{algorithmic}
\usepackage{graphicx}
\usepackage{textcomp}
\usepackage{wrapfig}
\usepackage{verbatim}
\usepackage{subfigure}
\usepackage{array}
\def\BibTeX{{\rm B\kern-.05em{\sc i\kern-.025em b}\kern-.08em
    T\kern-.1667em\lower.7ex\hbox{E}\kern-.125emX}}
\definecolor{abstractbg}{rgb}{0.89804,0.94510,0.83137}
\begin{document}
\title{ReLoc-PDR: Visual Relocalization Enhanced Pedestrian Dead Reckoning via Graph Optimization
}
\author{Zongyang Chen, Xianfei Pan and Changhao Chen 
\thanks{This work is supported by National Natural Science Foundation of China (NSFC) under the Grant 62103427, 62073331, 62103430. (Zongyang Chen and Xianfei Pan are co-firstauthors.) (Corresponding author: Changhao Chen.)}
\thanks{The authors are with the College of Intelligence Science and Technology, National University of Defense Technology, Changsha 410073, China (e-mail: zongyangChen@163.com; afeipan@126.com;  changhao.chen66@outlook.com).}}

\IEEEtitleabstractindextext{%
\fcolorbox{abstractbg}{abstractbg}{%
\begin{minipage}{\textwidth}%
\begin{wrapfigure}[12]{c}{3.1in}%
\includegraphics[width=3in]{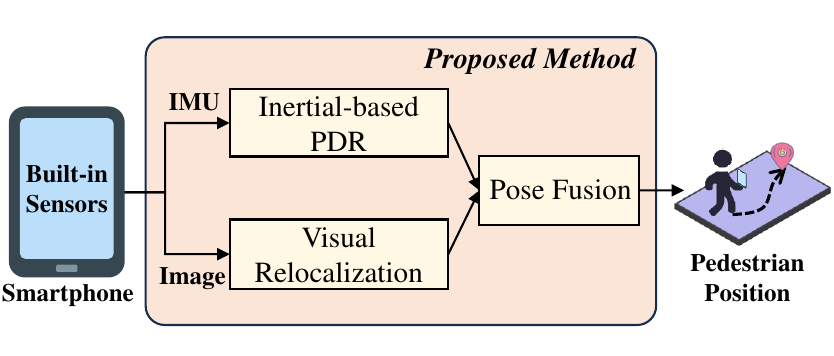}%
\end{wrapfigure}%
\begin{abstract}
Accurately and reliably positioning pedestrians in satellite-denied conditions remains a significant challenge. Pedestrian dead reckoning (PDR) is commonly employed to estimate pedestrian location using low-cost inertial sensor. However, PDR is susceptible to drift due to sensor noise, incorrect step detection, and inaccurate stride length estimation.
This work proposes ReLoc-PDR, a fusion framework combining PDR and visual relocalization using graph optimization. ReLoc-PDR leverages time-correlated visual observations and learned descriptors to achieve robust positioning in visually-degraded environments. A graph optimization-based fusion mechanism with the Tukey kernel effectively corrects cumulative errors and mitigates the impact of abnormal visual observations. 
Real-world experiments demonstrate that our ReLoc-PDR surpasses representative methods in accuracy and robustness, achieving accurte and robust pedestrian positioning results using only a smartphone in challenging environments such as less-textured corridors and dark nighttime scenarios. 
\end{abstract}

\end{minipage}}}

\maketitle

\section{Introduction}
\label{sec:introduction}
\IEEEPARstart{R}{obust} 
and accurate indoor pedestrian navigation plays a crucial role in enabling various location-based services (LBS). The determination of pedestrian location in satellite-denied environments is a fundamental requirement for numerous applications, including emergency rescue operations, path guidance systems, and augmented reality experiences~\cite{Smartphone-review, Emergency-responders}.
The existing indoor positioning solutions relying on the deployment of dedicated infrastructures are susceptible to signal interference and non-line-of-sight (NLOS) conditions, and their widespread deployment can be prohibitively expensive.

In contrast to infrastructure-based positioning methods, pedestrian dead reckoning (PDR) relies on only inertial data to estimate pedestrian location, providing a relatively robust approach in various environments. However, due to the inherent noise in inertial sensors, PDR is susceptible to trajectory drift over long-term positioning. To mitigate this issue, researchers have explored the combination of PDR with additional positioning methods, such as Ultra-Wideband (UWB), WiFi, Bluetooth, among others~\cite{PDR-UWB, PDR-WIFI, PDR-BLE}. These methods have demonstrated impressive results in indoor localization. However, they often necessitate the presence of extra infrastructure and require rigorous calibration procedures in advance.

The pursuit of a low-cost, robust, and self-contained positioning system is of great importance to flexible and resiliant pedestrian navigation. Visual relocalization, which estimates the 6 degree-of-freedom (DoF) pose of a query image against an existing 3D map model, holds potential for achieving drift-free global localization using only a camera. Considering that smartphones commonly are with built-in cameras, utilizing image-based localization methods on mobile devices becomes a viable approach. However, the challenge lies in the susceptibility of image-based relocalization to environmental factors such as changes in lighting conditions and scene dynamics.
Existing 3D structure-based visual localization methods~\cite{Prioritized-Matching2017Sattler, City-scale, HF-Net, InLoc} are computationally demanding and fail to provide real-time pose output. Furthermore, the positioning accuracy of visual relocalization significantly deteriorates in challenging environments due to the scarcity of recognizable features.

Considering the continuity and autonomy advantages of PDR, combining visual relocalization with PDR serves as a complementary approach. This integration allows for the correction of accumulated errors in PDR using visual relocalization results, while PDR improves the continuity and real-time performance of trajectory estimation. Several recent studies have begun exploring this direction~\cite{SFM-PDR, V-PDR, MEMS-Vision, MEMS-aid-IBL}. Existing methods primarily employ dynamic weighting strategies to loosely integrate PDR with visual relocalization. However, these approaches may lack robustness in challenging environments and can result in significant trajectory inconsistencies due to the interference caused by abnormal visual observations.


To tackle these challenges, this work presents ReLoc-PDR, a robust framework for pedestrian inertial positioning aided by visual relocalization. ReLoc-PDR leverages recent advancements in deep learning-based feature extraction \cite{SuperPoint} and graph optimization \cite{iSAM2Kaess2011} to ensure reliable visual feature matching and robust localization. By integrating these techniques, our method effectively mitigates the risk of visual relocalization failure, enhancing the system's robustness in visually degraded environments.
To fuse the pose results from PDR and visual relocalization effectively, we design a graph optimization-based fusion mechanism using the Tukey kernel. This mechanism facilitates cumulative error correction and eliminates the impact of abnormal visual observations on positioning accuracy. As a result, the ReLoc-PDR system exhibits stability and reliability. Real-world experiments were conducted to evaluate the performance of the proposed method. The results demonstrate the efficacy of our approach in various challenging environments, including texture-less areas and nighttime outdoor scenarios.

Our contributions can be summarized as follows:
\begin{itemize}
    \item We propose ReLoc-PDR, a robust pedestrian positioning system that effectively integrates Pedestrian Dead Reckoning (PDR) and visual relocalization to mitigate positioning drifts. 
    \item We design a robust visual relocalization pipeline that leverages learned global descriptors for image retrieval and learned local feature matching. It enhances the robustness of the positioning system, particularly in visually degraded scenes where traditional methods may struggle.
    \item We introduce a pose fusion mechanism by incorporating the Tukey kernel into graph optimization that facilitates cumulative error correction. It effectively eliminates the impact of abnormal visual observations on positioning accuracy, ensuring the stability and reliability of the ReLoc-PDR system.
\end{itemize}

\begin{figure*}
    \centering
    \includegraphics[width=0.95\hsize]{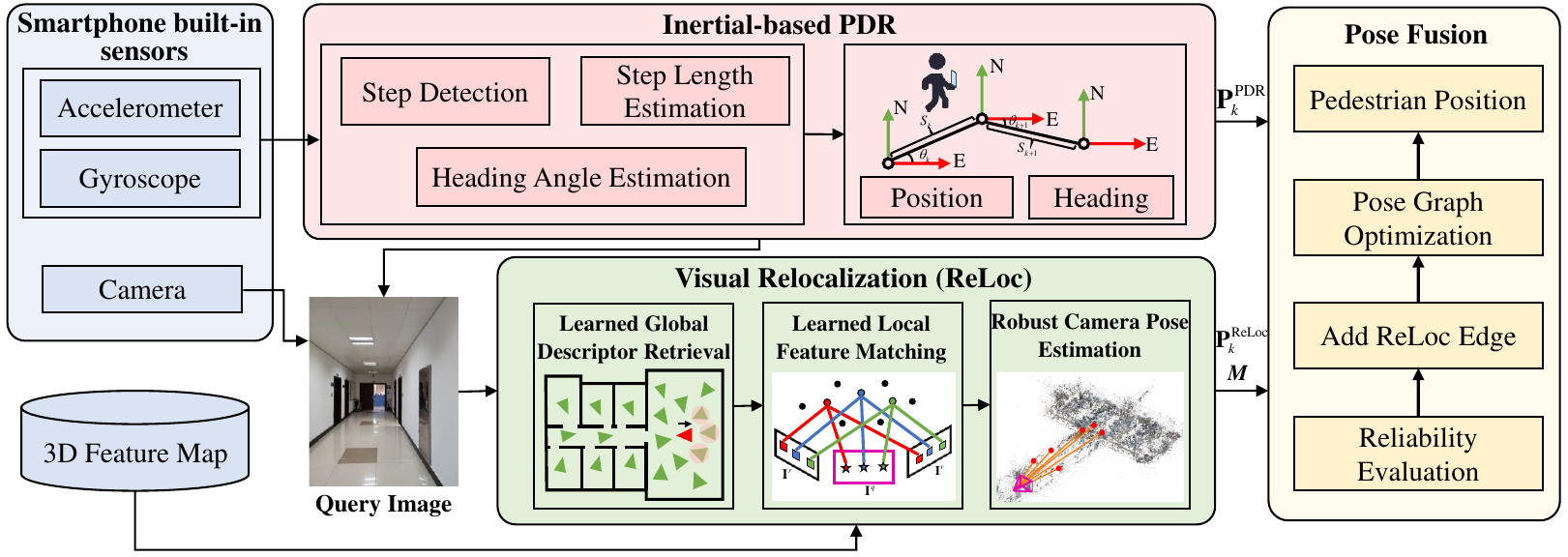}
    \caption{The system architecture of the proposed visual relocalization-aided PDR method.}
    \label{fig1}
\end{figure*}

\section{Related Work}
\subsection{Pedestrian Dead Reckoning}
Pedestrian dead reckoning (PDR) relies on measurements obtained from the built-in sensors of smartphones to detect human gait events and estimate step length and heading angle for pedestrian positioning~\cite{Smartphone-review, PDR}. However, PDR alone is prone to cumulative error over time, resulting in inaccurate positioning, particularly over long distances.
To address this limitation, previous studies have attempted to integrate PDR with other absolute localization technologies such as Wi-Fi, Bluetooth, and Ultra-Wideband (UWB). These integration approaches aim to periodically correct the accumulated error of PDR by incorporating absolute position information~\cite{PDR-UWB, PDR-WIFI, PDR-BLE}.
Despite their potential benefits, these methods are heavily reliant on infrastructure availability and can be susceptible to changes in the physical environment. Factors such as variations in signal strength, infrastructure coverage, and environmental conditions may impact the accuracy and reliability of these infrastructure-dependent approaches.

The integration of pedestrian dead reckoning (PDR) with visual relocalization has emerged as a promising research direction for achieving self-contained and highly accurate positioning using smartphones, without the need for additional infrastructure. This research area has gained increasing attention in recent years. Elloumi et al.~\cite{Comparion-Inertial-Vision} conducted a comparative study between inertial and vision-based methods for indoor pedestrian localization using smartphones. However, they did not explore the integration of these approaches to further enhance the performance of the localization system.
In another study, a refined indoor localization framework was proposed in~\cite{SFM-PDR}, which combines image-based localization with PDR to improve pedestrian trajectory estimation.
Wang et al.~\cite{VIO-PDR} introduced a vision-aided PDR localization system that integrates visual and PDR measurements into a unified graph model.
Furthermore, a novel indoor localization method called V-PDR was proposed in~\cite{V-PDR}, which integrates image retrieval and PDR using a weighted average strategy. This approach successfully reduces the accumulated error of PDR.
Shu et al.~\cite{MEMS-Vision} employed a dynamic fusion strategy that integrates PDR and image-based localization (IBL) based on the number of inliers in the IBL process. Their approach enables continuous and accurate 3D location estimation for long-term tracking using smartphones.
Additionally, in \cite{MEMS-aid-IBL}, a multimodal fusion algorithm was proposed that loosely couples PDR with visual localization to correct cumulative errors in PDR results.
However, despite these advancements, the robustness of these approaches remains limited, and the positioning accuracy may significantly degrade in visually challenging environments. Further improvements are necessary to ensure reliable and accurate positioning in such scenarios.

\subsection{Visual Relocalization}
Visual relocalization, also known as image-based localization, refers to the task of estimating the precise 6 DOF camera pose of a query image within a known map. This task can be approached using retrieval-based and structure-based methods.
Retrieval-based approaches~\cite{DBOW2, VLAD, NetVLAD, VPS-survery} estimate the pose of the query image by leveraging the geolocation of the most similar image retrieved from an image database. However, these methods often fall short in terms of localization accuracy.
Structure-based methods~\cite{Prioritized-Matching2017Sattler, City-scale, HF-Net, InLoc, SFM-PDR}, rely on establishing correspondences between 2D features in the query image and 3D points in a Structure from Motion (SFM) model using local descriptors. To handle large-scale scenes, these methods typically employ image retrieval as an initial step, restricting the 2D-3D matching to the visible portion of the query image~\cite{HF-Net, InLoc}. However, the robustness of traditional localization methods is limited due to the insufficient invariance of handcrafted local features.
In recent years, CNN-based local features~\cite{LF-Net, SuperPoint, D2-Net} have exhibited impressive robustness against illumination variations and viewpoint changes. These features have been employed to enhance localization accuracy.
For example, \cite{HF-Net} presents a comprehensive pipeline for structure-based visual relocalization, incorporating global image retrieval, local feature matching, and pose estimation. However, this method may encounter challenges during the image retrieval stage in scenes with significant appearance variations, as the representation power of global features is limited. Additionally, these methods often yield a low-frequency global position estimation and require high-performance computing platforms.

\section{Visual ReLocalization Enhanced Pedestrian Dead Reckoning}

\subsection{System Design}
\label{PDR}
\subsubsection{System Overview}
The framework of our ReLoc-PDR is depicted in Figure \ref{fig1}, comprising three primary modules: inertial sensor-based Pedestrian Dead Reckoning (PDR), visual relocalization, and graph optimization-based pose fusion. 
In this architecture, the PDR algorithm is employed to compute the per-step pose using the built-in inertial sensors of a commercially available smartphone.
The visual relocalization pipeline aims to accurately and robustly estimate the pose of the triggered image captured by the smartphone camera. This estimation is performed in relation to a pre-built 3D feature map, enabling global visual observations that periodically correct the accumulated error in the PDR.
Finally, the pose fusion module integrates the  pose results from PDR and visual relocalization using graph optimization with a robust Tukey kernel. This integration enables the continuous and smooth estimation of the pedestrian's position and trajectory during long-term walking.

\subsubsection{Pedestrian Dead Reckoning}
The PDR algorithm utilizes inertial data to estimate the pedestrian's position based on human motion characteristics. It consists of four main steps: step detection, step length estimation, heading angle estimation, and position update.
Step detection relies on the repetitive pattern observed in the accelerometer measurements during human walking. In our work, we employ a multi-threshold peak detection algorithm to identify pedestrian gait:
\begin{equation}
    \label{eq1}
    \left\{ \begin{array}{ll}
        \delta_{min} < |a_{m}-g| < \delta_{max} \\
        \Delta T >  \delta_{t}
    \end{array}
    \right.
\end{equation}
Here, $a_{m}$ represents the magnitude of acceleration, $\delta_{min}$ and $\delta_{max}$ denote the minimum and maximum acceleration values within one step, $g$ represents the gravity value, $\Delta T$ denotes the time interval between adjacent peaks, and $\delta_{t}$ represents the minimum duration threshold.
Considering the complexity of pedestrian movement and the presence of inevitable noises in low-cost MEMS inertial sensors, we employ a fourth-order low-pass filter to preprocess the acceleration signal. This preprocessing step enhances the quality of the gait characteristics obtained. Furthermore, to eliminate false peaks caused by external interference, we evaluate whether each peak is a local maximum value within a specific sliding window size. This additional criterion helps ensure the accuracy of the detected gait peaks.

The step length estimation component aims to calculate the distance covered by a pedestrian in a single step, which is influenced by the pedestrian's motion states. In our approach, we employ the Weinberg model~\cite{weinberg2002}  to estimate the pedestrian step length via:
\begin{equation}
\label{eq2}
S_{k} = K \cdot \sqrt[4]{a_{z,max}-a_{z,min}}
\end{equation}
Here, $K$ represents the calibrated step-length coefficient, while $a_{z,max}$ and $a_{z,min}$ denote the maximum and minimum values of vertical acceleration during step $k$, respectively.

Heading estimation is utilized to determine the walking direction of the pedestrian. We leverage the gyroscope data from the smartphone's built-in Inertial Measurement Unit (IMU) to estimate the pedestrian's heading angle. This is achieved through an attitude update equation based on the median integration method.
Finally, the pedestrian's position is updated based on their previous position, incorporating the estimated step length and heading angle. The updated position is as follows:
\begin{equation}
\label{eq3}
\left[\begin{matrix}
x_{k} \ y_{k}
\end{matrix}\right] = \left[\begin{matrix}
x_{k-1} \ y_{k-1}
\end{matrix}\right] + \left[\begin{matrix}
S_{k} \cos(\psi_{k}) \ S_{k} \sin(\psi_{k})
\end{matrix}\right]
\end{equation}
Here, $\psi_{k}$ represents the heading angle, while $x_{k}$ and $y_{k}$ indicate the pedestrian's horizontal position at step $k$.

\begin{figure*}[t]
    \centering
    \subfigure[Feature extraction]{
    \centering 
    \includegraphics[width=0.35\hsize]{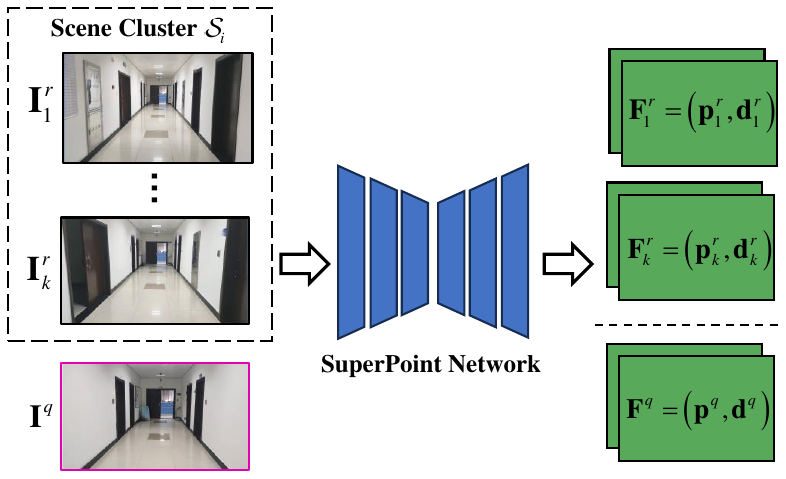}\label{fig2-a-feature-extraction}
    }\hspace{.5in}
    \subfigure[2D-3D correspondence]{
    \centering
    \includegraphics[width=0.3\hsize]{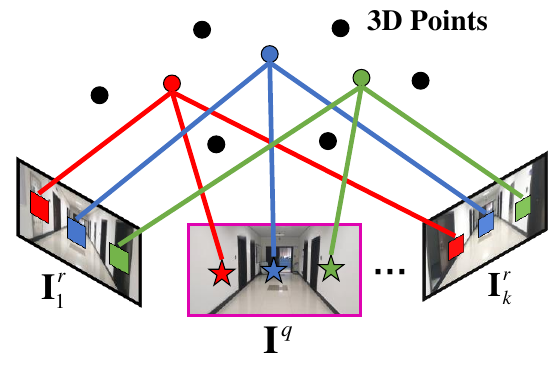}\label{fig2-b-feature-matching}
    }
    \caption{Local feature matching with learned feature via SuperPoint network. (a) KeyPoints and descriptors extraction for query image and each reference image in the scene cluster through SuperPoint network. (b) Establishing 2D-3D correspondence between 2D features detected in the query image and 3D points contained in the scene cluster depending on learned local feature matching.}
    \label{fig2-feature-matching}
\end{figure*}

\subsubsection{Data Flow}
As depicted in Figure~\ref{fig1}, the ReLoc-PDR system takes in acceleration and gyroscope data from the built-in inertial sensor of the smartphone to execute the PDR algorithm on the mobile device, incrementally calculating the pedestrian's position $\mathbf{P}_{k}^{\text{PDR}}$ based on human motion characteristics.

Once a step is successfully detected, a query image is triggered by capturing a photo using the smartphone's camera. Simultaneously, the captured query image, along with the current pedestrian pose, is transmitted to the visual relocalization module. Relocalization is achieved by matching the current image against a pre-constructed 3D sparse feature map and estimating the 6-DOF pose of the query image. Upon completing the visual relocalization pipeline, the ReLocalization module returns the pose of the query image $\mathbf{P}_{k}^{\text{ReLoc}}$ and the number of matched inliers $M$ to the pose fusion module.
Finally, the fusion module processes the results from the reLocalization and PDR module, and performs pose graph optimization in a separate thread.

\subsection{Visual Relocalization with Learned Feature}
Given that pedestrian navigation spans diverse indoor and outdoor environments, the visual relocalization method needs to exhibit robustness in handling various viewpoint and lighting conditions, including illumination, weather, seasonal changes, and consistent performance in both indoor and outdoor settings.
The robustness of traditional retrieval-based~\cite{DBOW2, VLAD} or structure-based relocalization~\cite{City-scale, Prioritized-Matching2017Sattler, SFM-PDR, MEMS-aid-IBL} methods is limited due to the insufficient invariance of handcrafted designed features~\cite{SIFT, ORB}, resulting in reduced stability performance under conditions with low texture or poor lighting.
Recent advancements in deep neural network-based methods, such as NetVLAD~\cite{NetVLAD} and SuperPoint~\cite{SuperPoint}, have demonstrated superior capabilities in image feature extraction, keypoint detection, and matching. These deep learning-based approaches surpass traditional baselines like bag-of-words~\cite{DBOW2}, VLAD~\cite{VLAD}, and SIFT~\cite{SIFT} in terms of robustness.
Motivated by these developments, we incorporate the advancements in learned global descriptors and learned local features into the visual relocalization pipeline. This integration enhances the robustness of the pedestrian positioning system in visually degraded scenarios.

The flowchart illustrating the visual relocalization process with learned features is presented in Figure \ref{fig1}.
Given a pre-built 3D sparse feature map database comprising images with known poses $\{\mathbf{I}^r_i, \mathbf{T}^r_i\}$, 3D point clouds $\{\mathbf{P}_j\}$, and global feature descriptors $\{f(\mathbf{I}^r_i)\}$, the objective of visual relocalization is to estimate the 6-DoF pose $\mathbf{T}^q$ of the query image $\mathbf{I}^q$ requested by the mobile device.
To accomplish this, we propose a hierarchical three-stage visual localization pipeline. Firstly, to identify similar reference images in the database, we employ a learned global descriptor~\cite{NetVLAD} and exploit co-visible information encoded in the sparse feature model to retrieve the top $N$ scene clusters $\mathcal{S}_{covis}$.
Next, we establish 2D-3D correspondences between the query image $\mathbf{I}^q$ and the 3D features ${\mathbf{P}_j}$ by leveraging 2D-2D matching with the learned Superpoint feature~\cite{SuperPoint}.
Finally, we solve for the camera pose $\mathbf{T}^q$ using the PnP~\cite{P3P} algorithm with a geometric consistency check within a RANSAC~\cite{RANSAC} loop.

\begin{figure*}
  \centering
  \includegraphics[width=0.85\hsize]{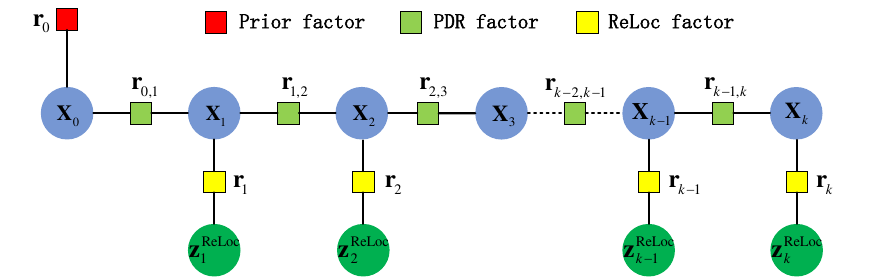}
  \caption{An illustration of the inertial-centric pose graph for fusing PDR and visual relocalization results. The pose graph include two types of edges, where the PDR edge connect the every two adjacent step nodes and is continuously added into graph with pedestrian steps. However, not every state node has the ReLoc edge, due to visual relocalization may fail or invalid in challenging scenarios.
  }
  \label{fig3}
\end{figure*}

\subsubsection{Learned Global Descriptor based Image Retrieval}
Image retrieval aims to identify a subset of database images that share a covisibility relationship with the query image.
For the query image $\mathbf{I}^q$, we initially extract learned global descriptors $f(\mathbf{I}^q)$ using the NetVLAD model (Remarkable CNN-based descriptor)~\cite{NetVLAD}, which has demonstrated superior performance compared to non-learned image representations~\cite{DBOW2, VLAD} in visual place recognition benchmarks.
Subsequently, we retrieve the top $k$ reference images $\mathcal{\tilde{S}}={\tilde{\mathbf{I}}^r_1, \tilde{\mathbf{I}}^r_2, \cdots, \tilde{\mathbf{I}}^r_k}$ from the sparse features model based on the distance metric $d(f(\mathbf{I}^q), f(\mathbf{I}^r))$.
Rather than directly clustering these retrieved reference images based on co-observed 3D points or performing feature matching between the query image and retrieved reference images (as done in HF-Net)~\cite{HF-Net}, we expand the retrieval results by leveraging the co-visible information embedded in the sparse features model. This approach helps to search for additional potential 3D points and enlarge the view region.
Specifically, for each retrieved image $\tilde{\mathbf{I}}^r_i$, we search for the $n$ database images in the Structure from Motion (SFM) model that share the same observed 3D map points. We then merge these images to form a scene cluster $\mathcal{S}_i$, with the size of each scene cluster limited to 20 images to reduce computational costs.
If a retrieved image $\tilde{\mathbf{I}}^r_i$ already exists in a previous cluster, it is disregarded to avoid redundant expansion of the retrieval results.
Finally, we obtain $N$ scene clusters $\mathcal{S}_{covis} = { \mathcal{S}_1, \cdots, \mathcal{S}_N }$ and sort them according to the size of the cluster set.

\subsubsection{Learned Local Feature Matching}
After obtaining the co-visible candidates $\mathcal{S}_{covis}$ for the query image $\mathbf{I}^q$, our focus shifts to establishing 2D-3D correspondences between the query image and 3D points based on 2D-2D matching utilizing the SuperPoint learned feature~\cite{SuperPoint}. SuperPoint is a self-supervised framework that extracts both interest points and descriptors. It has been demonstrated to generate denser and more accurate matching results compared to traditional handcrafted detectors like SIFT~\cite{SIFT} and ORB~\cite{ORB}. Moreover, SuperPoint exhibits excellent robustness against illumination changes.
For each scene cluster $\mathcal{S}_i \in \mathcal{S}_{covis}$, we begin by extracting the set of ketpoint positions $\mathbf{p}^q$ and associated local descriptors $\mathbf{d}^q$ of the query image using the SuperPoint network, as illustrated in Figure~\ref{fig2-a-feature-extraction}, jointing them $\left( \mathbf{p}^q, \mathbf{d}^q \right)$ as local features $\mathbf{F}^q$. Meanwhile, the 2D features of reference images, represented by $\mathbf{F}^r_i=\left(\mathbf{p}^r_i, \mathbf{d}^r_i \right)$, within the cluster set $\mathcal{S}_i$.

During the process of local feature matching, we employ nearest neighbor (NN) matcher to establish 2D-2D matches between the query features $\mathbf{F}^q$ and each reference feature $\mathbf{F}^r_i$ through visual descriptors similarity, followed by a Lowe’s ratio test~\cite{lowe-ratio} to filter out ambiguous matched pairs.
Finally, the 2D-3D correspondences are established by retrieving the 3D points corresponding to the keypoints detected in the database image based on the previous 2D-2D match results (Figure~\ref{fig2-b-feature-matching}). These correspondences are crucial for solving the camera pose estimation. 

\subsubsection{Robust Pose Estimation}
For the query image $\mathbf{I}^q$, we effectively establish the 2D-3D correspondence between the 2D keypoints detected in the query image and the 3D points within the scene cluster $\mathcal{S}_i$. Upon building this correspondence, we proceed to solve the pose of the query image using the PnP algorithm~\cite{P3P} within a RANSAC loop~\cite{RANSAC}.
During the iterative solving process, if the number of inliers in the estimated pose surpasses a predefined threshold, the program will terminate early. This early termination condition ensures efficiency and expedites the pose estimation procedure.

\subsection{Integrating PDR and Visual Relocalization via Factor Graph Optimization}
In the process of visual and inertial pose fusion, ensuring the reliability of observation information is crucial for maintaining the stability of the multi-sensors positioning system.
However, we observed that evaluating the quality of visual relocalization solely based on the number of inliers~\cite{MEMS-Vision, MEMS-aid-IBL} is not robust enough in visually degraded scenes. Examples of such scenes include texture-less walls, areas with similar structures, and dark roadways. In these scenarios, abnormal visual observations can still be introduced into the positioning system, leading to a significant degradation in accuracy.

To address this challenge, we propose a robust pose fusion algorithm that integrates Pedestrian Dead Reckoning (PDR) with visual relocalization using graph optimization~\cite{iSAM2Kaess2011} and the Tukey robust kernel~\cite{Turkey-kernel}. By incorporating the Tukey kernel function into the pose graph, we can adaptively assess the impact of current visual relocalization results on the system's states. This adaptive assessment dynamically determines the weight of the visual observation, effectively mitigating the risk of visual relocalization failures.
Furthermore, unlike existing visual-inertial fusion methods~\cite{VINS-Mono}, we employ an inertial-centric data processing scheme. This scheme enables the dynamic integration of visual relocalization observations into the graph. In the pose graph, as illustrated in Figure~\ref{fig3}, each step node serves as a vertex, connecting to other vertices through two types of edges.

\begin{figure*}
  \centering
  \subfigure[Indoor 3D feature map]{
  \centering 
  \includegraphics[height=0.30\hsize]{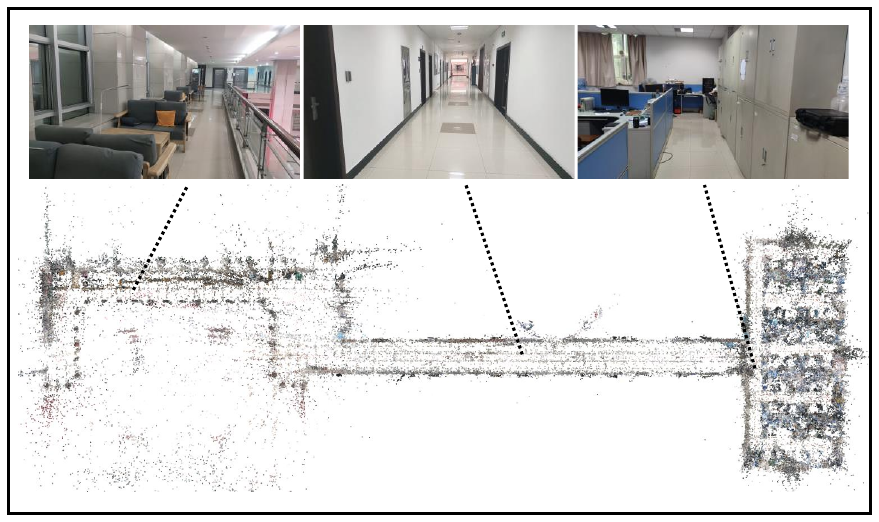}\label{fig4-a-indoor-map}
  }~\subfigure[Outdoor 3D feature map]{
  \centering
  \includegraphics[height=0.30\hsize]{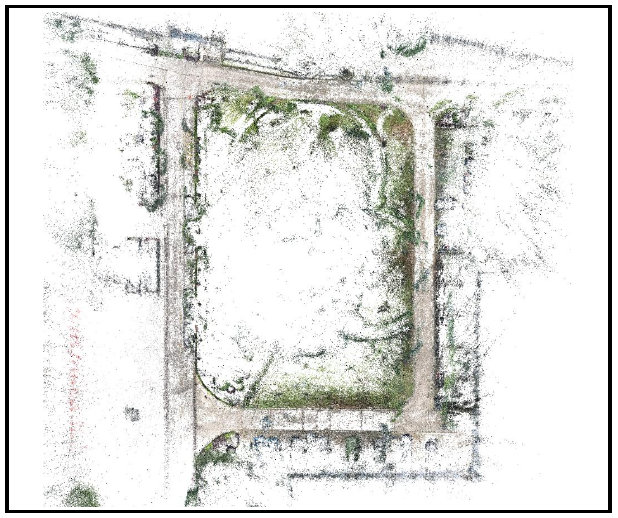}\label{fig4-b-outdoor-map}
  }
  \caption{The reconstruction results of 3D feature map with SuperPoint features. (a) Indoor experimental environment map. (b) Outdoor experimental environment map.}
  \label{fig4}
\end{figure*}

\subsubsection{PDR Factor}
As discussed in Section \ref{PDR}, the Pedestrian Dead Reckoning (PDR) algorithm offers the advantages of autonomy and continuity, enabling high-accuracy positioning within a short period.
Upon successful detection of a step during pedestrian walking, a PDR factor is established to connect it with the previous step. This PDR factor represents the relative change in the pedestrian's position and is obtained directly from the PDR algorithm.
Given our inertial-sensor-centric pose graph, it continuously expands as pedestrian steps are taken, rendering it relatively robust to environmental variations.
For step $k$ and its previous step $k-1$, the residual of the PDR factor is formulated as follows:
\begin{equation}
  \label{eq8}
  \begin{aligned}
    \mathbf{r}_{k,k-1}(\mathbf{z}_{k}^{\text{PDR}}, \mathbf{x}_{k-1}, \mathbf{x}_{k}) & = \mathbf{z}_{k}^{\text{PDR}} - f\left(\mathbf{x}_{k-1} , \mathbf{x}_{k}\right) \\
    & = \left[\begin{array}{c}
      S_{k} \cos (\psi_{k}) - (x_k-x_{k-1}) \\
      S_{k} \sin (\psi_{k}) -(y_k-y_{k-1})
    \end{array}
    \right]
    \end{aligned}
\end{equation}
Here, $\mathbf{r}_{k,k-1}(\cdot)$ represents the residual of the PDR factor. $\mathbf{z}_{k}^{\text{PDR}}$ denotes the PDR observation, which corresponds to the position increment from step $k$ to $k-1$. The state estimations at steps $k-1$ and $k$ are denoted as $\mathbf{x}_{k-1}$ and $\mathbf{x}_{k}$, respectively, after optimization.

\subsubsection{Relocalization Factor}
Before incorporating relocalization edges into the pose graph, a reliability assessment is performed to mitigate the impact of visual relocalization failures on positioning accuracy.
In our approach, we utilize the number of inliers as the criterion to determine the success of visual relocalization results. If the number of inliers exceeds 25, we consider the relocalization results reliable. In such cases, we add a ReLoc edge to the current state node and subsequently perform incremental smoothing optimization. However, if the number of inliers is below the threshold, we skip the pose graph optimization step and rely solely on the previously optimized state and PDR estimation to determine the current pedestrian position.

Assuming that the visual relocalization result is reliable at step $k$, the residual of the relocalization factor is calculated using equation \eqref{eq9}:
\begin{equation}
  \label{eq9}
  \begin{aligned}
    \mathbf{r}_{k}(\mathbf{z}_{k}^{\text{ReLoc}}, \mathbf{x}_{k}) & = \mathbf{z}_{k}^{\text{ReLoc}} - \mathbf{x}_{k} \\
    & = [\mathbf{T}_{k}^{\text{q}}]_{\mathbf{p}_{xy}} - \mathbf{x}_{k}
    \end{aligned}
\end{equation}
Here, $\mathbf{r}_{k}(\cdot)$ represents the residual of the relocalization factor. $\mathbf{z}_{k}^{\text{ReLoc}}$ denotes the visual relocalization observation, specifically the horizontal position $[\cdot]{\mathbf{p}_{xy}}$ of the camera pose. Additionally, the prior factor information $\mathbf{r}_{0}$ is also provided by the visual relocalization result, which aids in determining the initial position of the pedestrian before walking.

\subsubsection{Incremental Smoothing Optimization for Poses Fusion}
The set of states $\mathcal{X}$ comprises the state variables from the first step to the $k$th step, represented as $\mathcal{X} = \{\mathbf{x}_0, \mathbf{x}_1, \cdots, \mathbf{x}_k\}$.
The entire pose graph, consisting of PDR factors and ReLoc factors, is optimized by minimizing the following cost function:
\begin{equation}
  \label{eq10}
  {\mathcal{X}^*} = \mathop {\arg \min }\limits_{\mathcal{X}} \left\{ {\sum_{k \in \mathcal{B}}  {\left\| {\mathbf{r}_{k-1,k}} \right\|_{\mathbf{\Omega} _k^\text{PDR}}^2 
  +  \sum_{i \in \mathcal{L} } \rho \left( \left\| {\mathbf{r}_{i}} \right\|_{\mathbf{\Omega} _i^\text{ReLoc}}^2 \right) } } \right\}
\end{equation}
\begin{equation}
  \label{eq11}
  \rho(x) = \left\{\begin{array}{ll}
    \frac{1}{6} {\delta}^2 \left(1-\left[1-(x/{\delta})^{2}\right]^{3}\right), & \text{if} \left|{x}\right| \leq \delta \\
    \frac{1}{6} {\delta}^2, & \text{otherwise} 
    \end{array}\right.
\end{equation}
Here, $\mathcal{B}$ is the set of all PDR factors, $\mathcal{L}$ is the set of all ReLoc factors, and $\mathbf{\Omega}_k^{\text{PDR}}$ and $\mathbf{\Omega}_i^{\text{ReLoc}}$ denote the corresponding covariance matrices.
To account for variations in the quality of relocalization results under different visual conditions, we dynamically adjust the weight of the ReLoc edges based on the number of inliers.
Although the reliability evaluation strategy based on the number of inliers has helped eliminate incorrect visual relocalization results, we further incorporate the Tukey robust kernel $\rho(\cdot)$~\cite{Turkey-kernel} to mitigate the potential impact of any erroneous visual observations. The loss function of the Tukey robust kernel is defined in equation \eqref{eq11}.
In contrast, we do not employ any kernel function for the PDR edge, as PDR can achieve high-accuracy positioning within a short period.

To achieve real-time pose optimization, we utilize an adaptive-lag smoother called Incremental Smoothing and Mapping (iSAM2)~\cite{iSAM2Kaess2011}. Unlike batch optimizers that repeatedly compute and update all historical states, iSAM2 dynamically determines which historical states are affected by the current observations and selectively optimizes and updates only those affected states. This adaptive approach significantly reduces unnecessary computations, resulting in near-optimal results comparable to batch graph optimization but at a lower computational cost.
For implementation, we employ the open-source GTSAM library~\cite{Frank} to construct the factor graph and perform incremental smoothing optimization. The use of GTSAM enables efficient construction and manipulation of the factor graph, facilitating the real-time pose optimization process.

\section{Experiments}
In this section, we first present the experimental setup, encompassing the necessary equipment and intricate details of the experiment. Subsequently, we conduct comprehensive experiments to assess the robustness and accuracy of the fusion positioning system proposed in this study. We evaluate our method across three distinct environmental conditions, i.e. a textureless corridor, overcast weather, and a dark roadway, each presenting unique visual challenges. 

\subsection{Experimental Setup}
In the experiment, a Xiaomi 10 smartphone was employed for both offline map construction and online testing purposes. To reconstruct the 3D map model, video sequences of the scene were captured using the smartphone at a frequency of 30 Hz and a resolution of 1920x1080. These sequences were then downsampled to obtain discrete database images. In this study, COLMAP~\cite{COLMAP}, a structure-from-motion (SFM) tool, was utilized for generating sparse SFM models. We made certain modifications to adapt it to pedestrian navigation. Specifically, we employed the NetVLAD~\cite{NetVLAD} feature to retrieve the top 50 image matching pairs for each database image, which were subsequently inputted into the COLMAP pipeline to aid the image matching process. Additionally, we introduced a scale estimation module into COLMAP to convert the 3D point cloud into real-world scale, allowing for an understanding of the environment's dimensions. This approach involved using pre-placed artificial markers~\cite{AprilTag2} with known lengths to restore the 3D point cloud to a real-world scale. Finally, a new 3D SFM model was constructed using keypoints detected by SuperPoint~\cite{SuperPoint}, based on the hloc toolbox~\cite{HF-Net}. The resulting 3D SFM model, depicting both indoor and outdoor environments, can be observed in Fig.~\ref{fig4}.

During the testing phase, our system records IMU data at a rate of 100Hz and captures image frames at 30Hz. However, the query image is triggered specifically on the node where gait is detected during pedestrian walking. The resolution of all query images is standardized to 600x800 pixels. 
To assess the localization performance and robustness of the proposed method, we conducted experiments in three distinct environments characterized by varying visual challenges. These experiments involved holding a smartphone in each of these environments.

\subsection{Indoor Experiment in Textureless Corridor}

\begin{figure}[t]
  \centering
  \includegraphics[width=0.7\hsize]{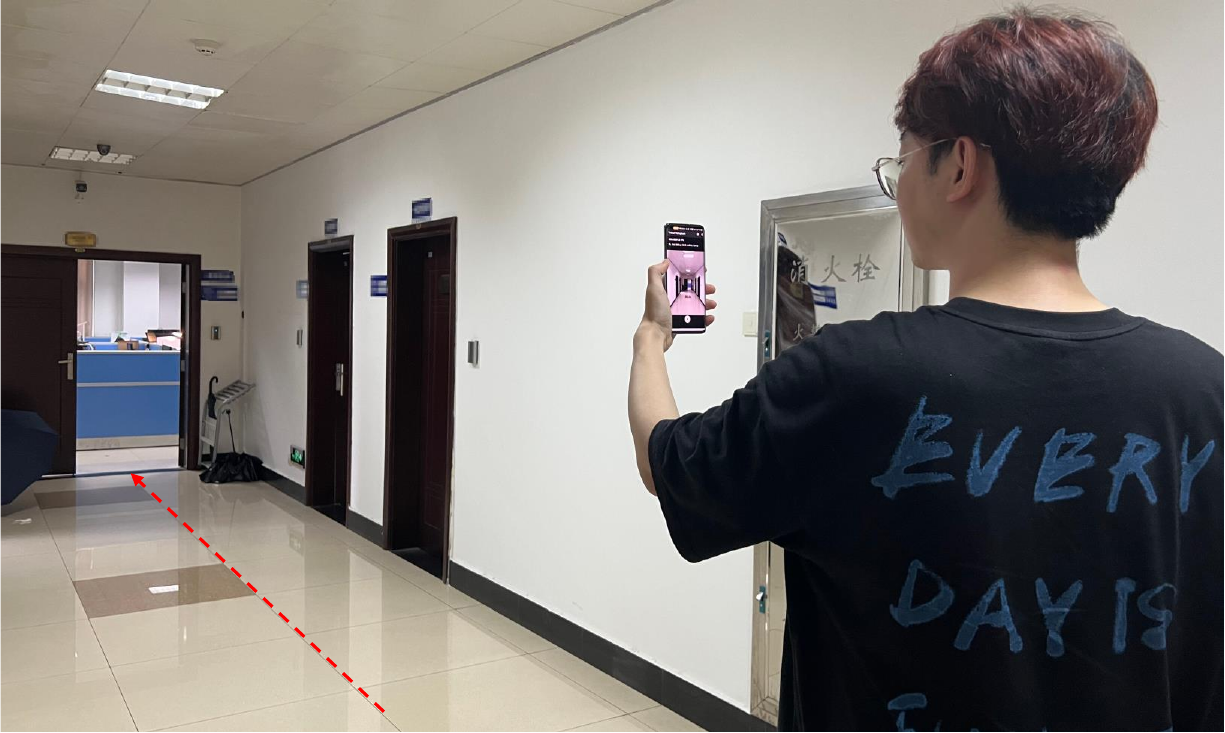}
  \caption{The process of indoor experiment. The volunteer walked along a preset route (red line) with holding a smartphone (Xiaomi 10).}
  \label{fig5}
\end{figure}

As illustrated in Fig.~\ref{fig5}, the initial experiment took place in an indoor corridor and office environment, known for its visually challenging characteristics such as walls with limited texture and the presence of moving pedestrians. During this experiment, the participant navigated the corridor while holding the smartphone and encountered multiple sharp turns along a pre-determined path. These sharp turns have the potential to induce significant heading drift. The experiment had a duration of approximately 355 seconds, covering a total distance of approximately 240 meters.

To highlight the advantages of our method in indoor positioning environments, we conducted a comparative analysis with three other approaches. The first method examined was a pure inertial-based pedestrian navigation (PDR) approach. The second approach involved VINS-Mono~\cite{VINS-Mono}, which is a state-of-the-art Visual-Inertial SLAM method known for its exceptional tracking performance and competitive positioning accuracy. The third approach combined PDR with visual localization using a dynamic weighting strategy~\cite{MEMS-Vision, MEMS-aid-IBL}, referred to as DW-PDR/vision.

\begin{figure}[t]
  \centering
  \includegraphics[width=0.8\hsize]{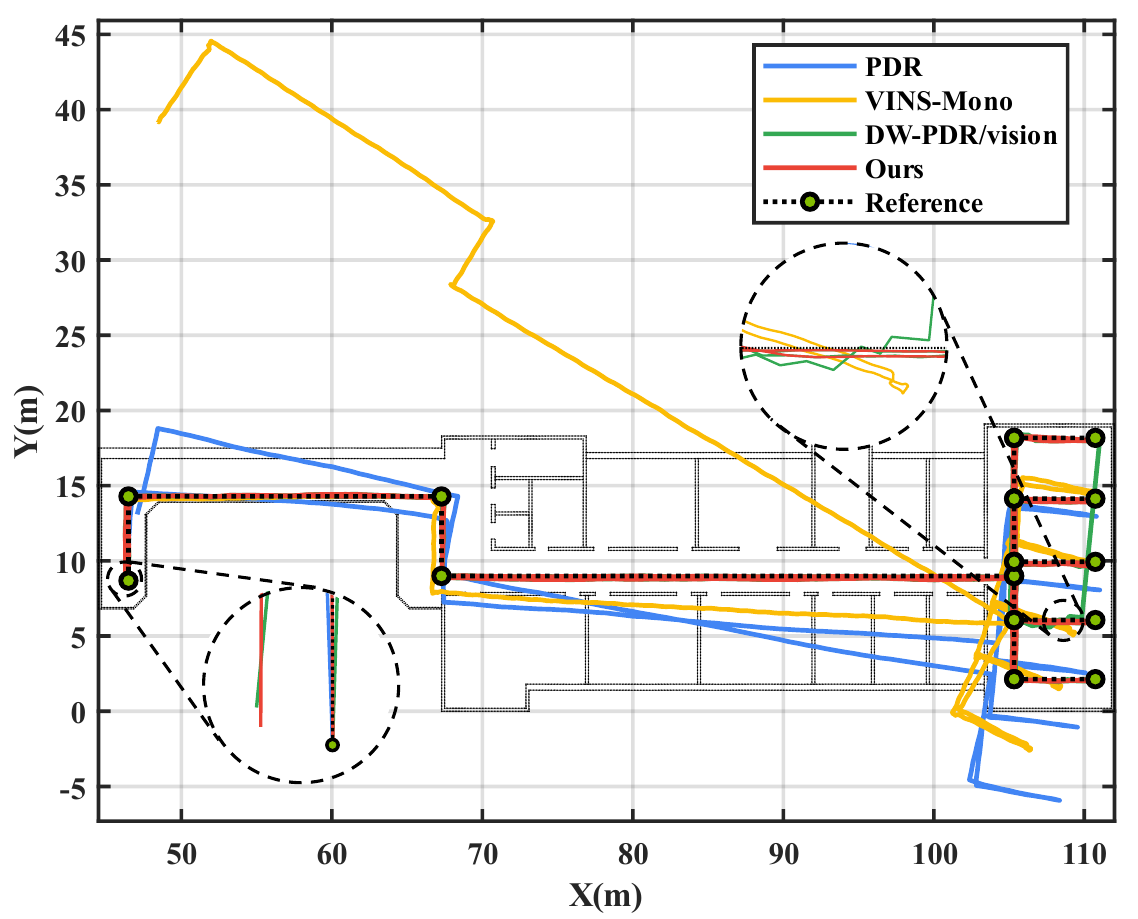}
  \caption{Comparison of the trajectory estimated by different methods in textureless corridor.}
  \label{fig6}
\end{figure}

\begin{table}[t]
  \renewcommand\arraystretch{1.25}
  \centering
  \caption{Horizontal positioning error statistics in indoor experiment}
  \label{tab1}
  \begin{tabular}{m{55pt}<{\centering}m{48pt}<{\centering}p{48pt}<{\centering}m{48pt}<{\centering}}
    \hline
     & RMSE(m) & Max Error(m) & Loop Error(m) \\ \hline
    PDR & 5.5951 & 8.4082 & 4.4419 \\ 
    VINS-Mono & 10.9209 & 30.7579 & 30.6570 \\ 
    DW-PDR/vision & 2.5575 & 12.4825 & 0.2274 \\ 
    Ours & \textbf{0.2079} & \textbf{0.4435} & \textbf{0.1955} \\ 
    \hline
  \end{tabular}
\end{table}

Figure~\ref{fig6} illustrates the trajectory comparison among the different methods in an indoor environment. The PDR algorithm demonstrates relatively smooth overall trajectory; however, it suffers from trajectory drift due to accumulated errors in heading estimation. The positioning accuracy of VINS-Mono is severely compromised in indoor environments due to the presence of less-textured features and the utilization of low-quality sensors in mobile devices, leading to inferior performance compared to the pure inertial-based PDR.
Leveraging stronger geometric constraints from a prior 3D map and the robustness of learned features, the visual relocalization method achieves accurate positioning in most cases. By combining PDR with the visual relocalization results, the cumulative errors in PDR can be effectively corrected using the visual measurements. However, it is evident that the trajectory of DW-PDR/vision lacks robustness and smoothness, often experiencing significant discontinuities due to interference from abnormal visual relocalization observations in visually similar scenarios.
In contrast, our proposed method exhibits robustness against abnormal visual relocalization observations. It dynamically assesses the reliability of visual relocalization results using the Tukey robust kernel, enabling adaptive decision-making regarding the reliance on either PDR or global visual observations. Additionally, our method leverages the incremental smoothing iSAM2 algorithm to provide a smoother and more continuous trajectory compared to other approaches, as depicted in the locally enlarged region in Figure~\ref{fig6}.

Table~\ref{tab1} provides the statistical analysis of horizontal positioning errors for the different methods. As obtaining the ground truth of pedestrian trajectory is not feasible, we adopt artificial marker points with known positions as a reference benchmark. The results presented in Table~\ref{tab1} indicate that our proposed method achieves superior positioning accuracy in complex indoor environments, effectively reducing the root mean square error (RMSE) of the pure inertial-based PDR by 96.3\%.
VINS-Mono exhibits the lowest accuracy due to its degraded tracking performance in indoor environments characterized by less-textured conditions. The DW-PDR/vision method experiences a maximum error of 12.4825 m, attributed to the influence of abnormal visual observations.
In comparison, our method surpasses the performance of DW-PDR/vision by improving the positioning accuracy by 91.9\% in terms of RMSE and reducing the maximum error to 0.4435 m. These results effectively demonstrate the robustness of our method in challenging indoor environments.

\begin{figure}[t]
  \centering
    \includegraphics[width=0.7\linewidth]{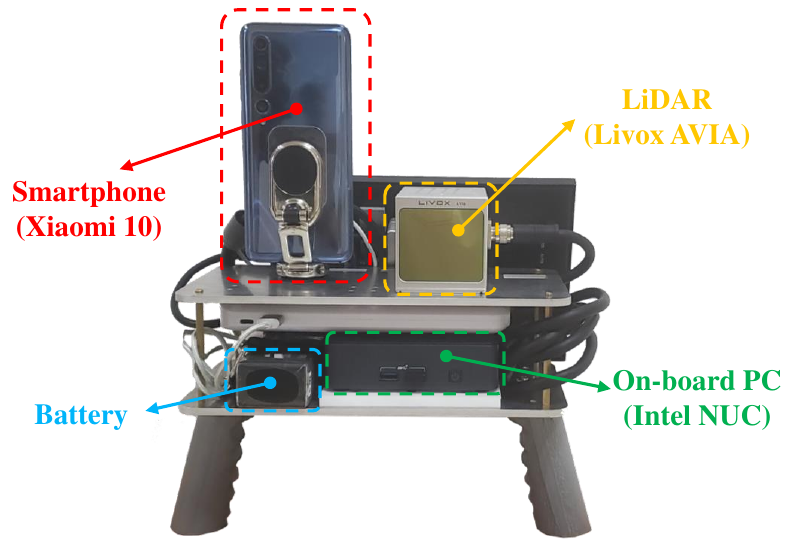}
  \caption{The handheld experimental equipment, integrating the smartphone (mi 10) and solid-state LiDAR (livox AVIA) on one platform.}
  \label{fig7}
\end{figure}

\begin{figure}[t]
  \centering
  \includegraphics[width=0.8\hsize]{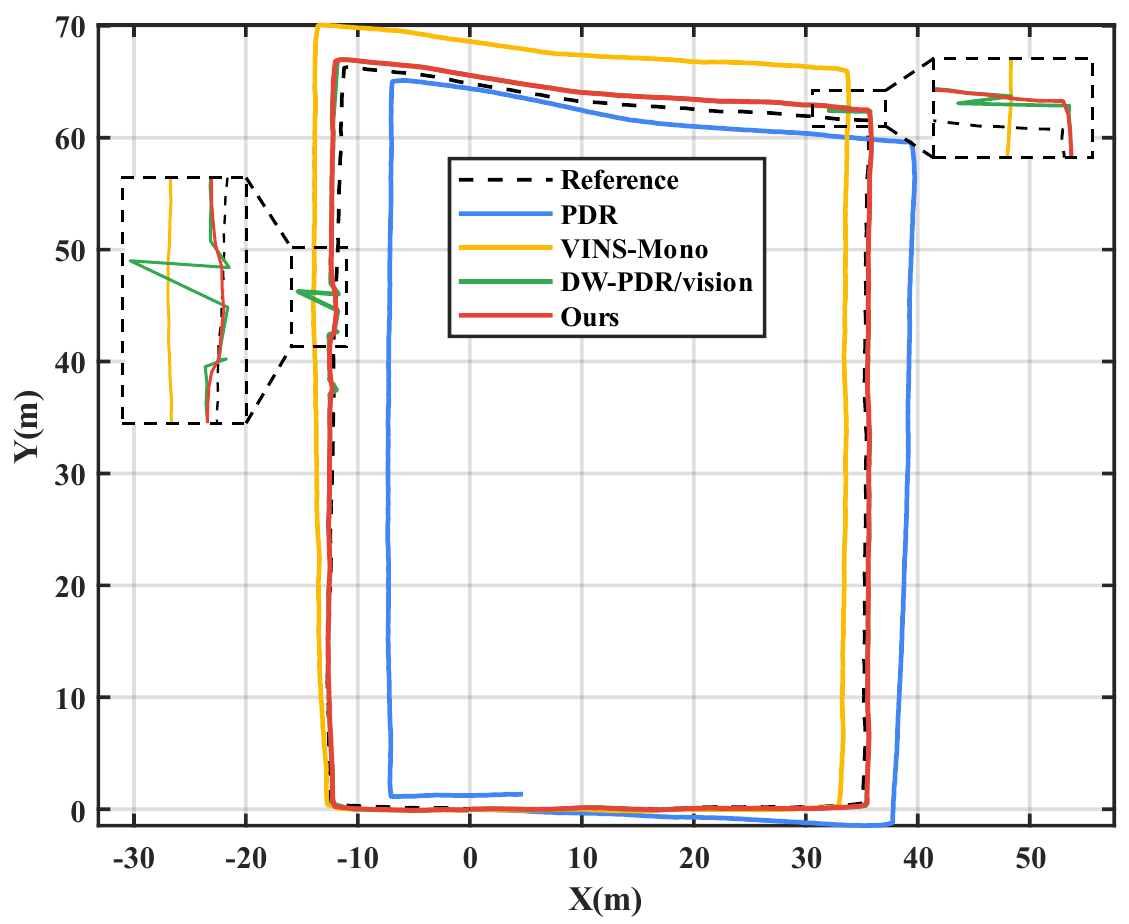}
  \caption{The trajectories estimated by different methods in outdoor overcast environments.}
  \label{fig8}
\end{figure}

\subsection{Outdoor Experiment in Overcast Weather Condition}
To evaluate the positioning performance of our proposed method in challenging outdoor environments, a second experiment was conducted along a route encircling a hill. This test encompassed dynamic vehicle movements, variations in cloudy weather, and changes in scene structure, all of which have the potential to impede visual tracking. The experiment had a walking duration of approximately 230 seconds, covering a total path length of approximately 225 meters.
As the satellite signal was obstructed by tall trees and buildings, it was not feasible to obtain a reference positioning trajectory from the RTK recorder. Instead, we employed the trajectory estimation of FAST-LIO2~\cite{FAST-LIO2}, one of the state-of-the-art LiDAR-inertial odometers, as a reference value. To synchronize the timestamps between the smartphone-based results and the LiDAR-based output, the volunteer held the experimental device (Fig.~\ref{fig7}) in an upward position to stimulate the accelerometer and produce a spike before walking. By aligning the first peak of the smartphone acceleration data with the first peak of the LiDAR's built-in IMU acceleration data, we obtained the time difference between them, achieving synchronization.

\begin{figure}[t]
  \centering
  \includegraphics[width=0.8\hsize]{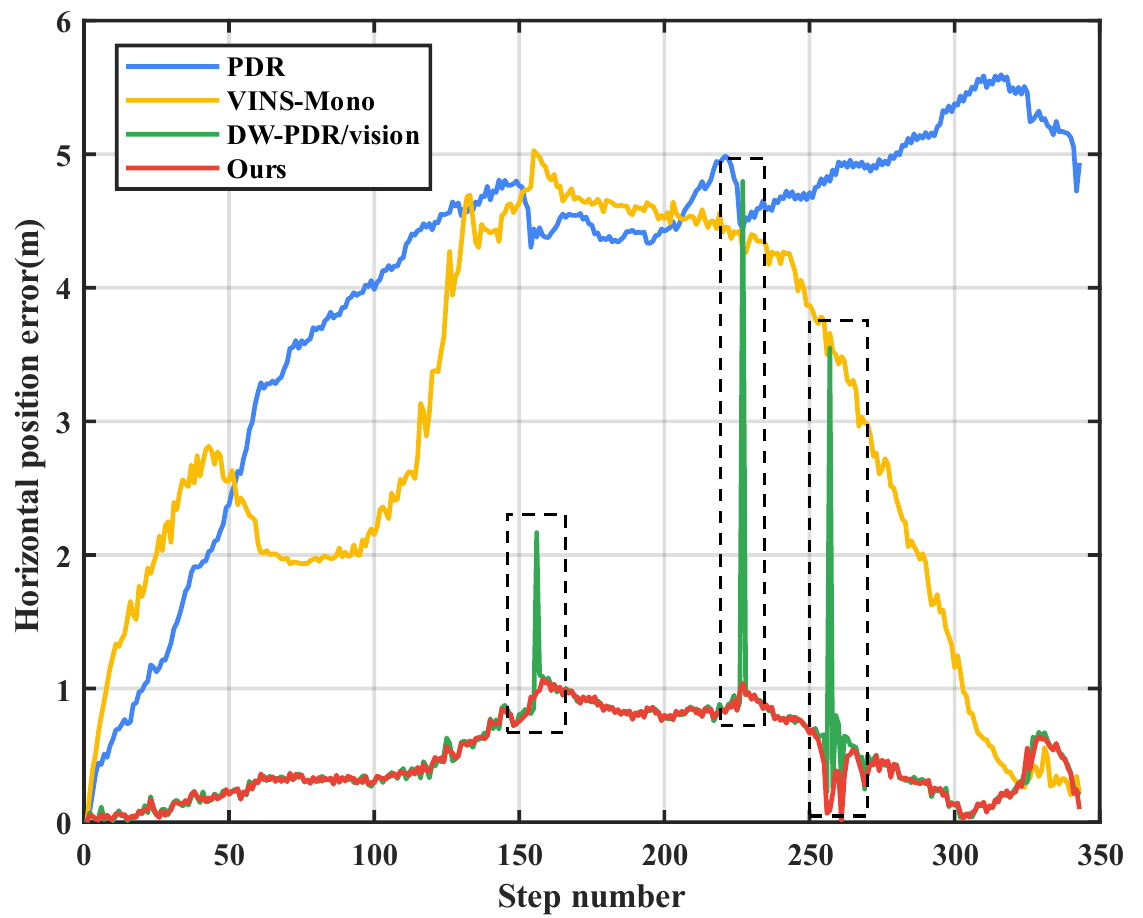}
  \caption{Distribution of horizontal position error comparison with the change of detected steps.}
  \label{fig9}
\end{figure}

\begin{table}[t]
  \renewcommand\arraystretch{1.25}
  \centering
  \caption{Comparison of positioning accuracy in outdoor overcast weather condition}
  \label{tab2}
  \begin{tabular}{p{55pt}<{\centering}p{48pt}<{\centering}p{48pt}<{\centering}p{48pt}<{\centering}}
    \hline
     & RMSE(m) & Max Error(m) & Loop Error(m) \\ \hline
    PDR & 4.2875 & 5.5949 & 4.9342 \\ 
    VINS-Mono & 3.2201 & 5.0276 & 0.2182 \\ 
    DW-PDR/vision & 0.6605 & 4.7965 & 0.2036 \\ 
    Ours & \textbf{0.5618} & \textbf{1.0664} & \textbf{0.0984} \\ 
    \hline
  \end{tabular}
\end{table}

Figure~\ref{fig8} displays the trajectory results of various algorithms in outdoor cloudy environments. It is evident that our proposed method closely aligns with the reference 
trajectory and exhibits a smooth trajectory without any sudden jumps. The PDR algorithm, due to the inherent noise of inertial sensors, significantly deviates from the ground truth. While VINS-Mono performs reasonably well in outdoor environments, its positioning accuracy is limited by the lower quality of mobile phone built-in sensors.
Compared to visual-inertial SLAM methods, the visual relocalization aided PDR methods achieve superior trajectory estimation. However, the DW-PDR/vision approach based on dynamic weighting strategy, despite achieving remarkable positioning results, demonstrates noticeable trajectory jumps under abnormal visual relocalization observations, a phenomenon not observed in our method.
The proposed optimization-based fusion positioning method effectively mitigates the impact of erroneous visual observations on positioning accuracy through the use of a robust kernel function, resulting in smoother and more robust positioning outcomes. Furthermore, the distribution of horizontal positioning errors with pedestrian steps is depicted in Figure~\ref{fig9}. These results highlight the superior performance of our method compared to other algorithms, consistently providing accurate positioning results with errors of less than 1 m. By incorporating the visual relocalization results, our method significantly reduces cumulative errors in PDR. In contrast, the positioning accuracy of the DW-PDR/vision method degrades significantly due to the interference of abnormal visual relocalization observations, underscoring the robustness of our method in challenging environments.

Table~\ref{tab2} presents the positioning error statistics for different algorithms. Our method exhibits the highest positioning accuracy and significantly reduces the cumulative errors of PDR by 86.9\%.
While VINS-Mono achieves impressive loop error reduction through loop-closing, its positioning accuracy does not match the competitiveness of our method.
In comparison to DW-PDR/vision, our method outperforms it by demonstrating a 14.9\% improvement in terms of the RMSE metric. Moreover, our method exhibits a maximum error of only 1.0604 m, whereas DW-PDR/vision shows a maximum error of 4.7975 m. These results effectively illustrate the robustness of our proposed method against abnormal disturbances.

\begin{figure}[t]
  \centering
  \includegraphics[width=\hsize]{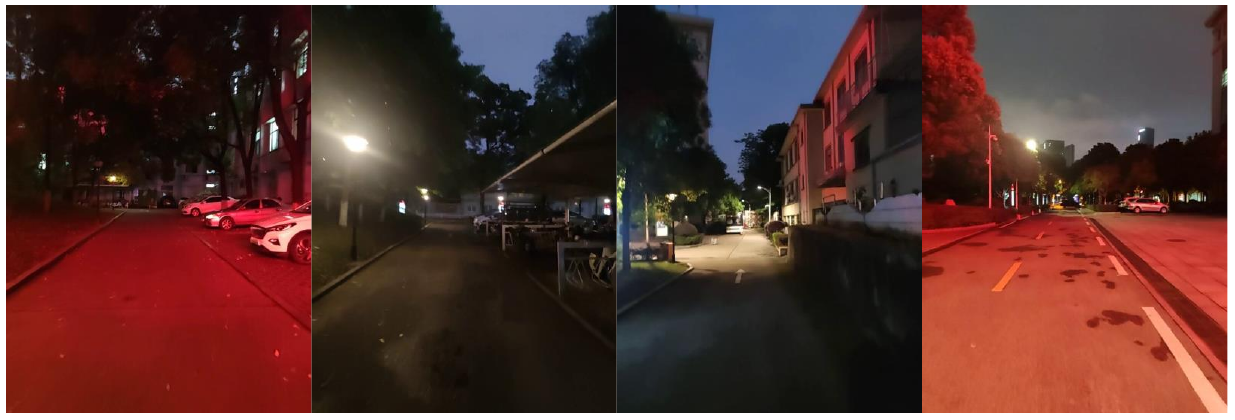}
  \caption{Example query images for outdoor nighttime experiment.}
  \label{fig10}
\end{figure}

\begin{figure}[t]
  \centering
  \includegraphics[width=0.8\hsize]{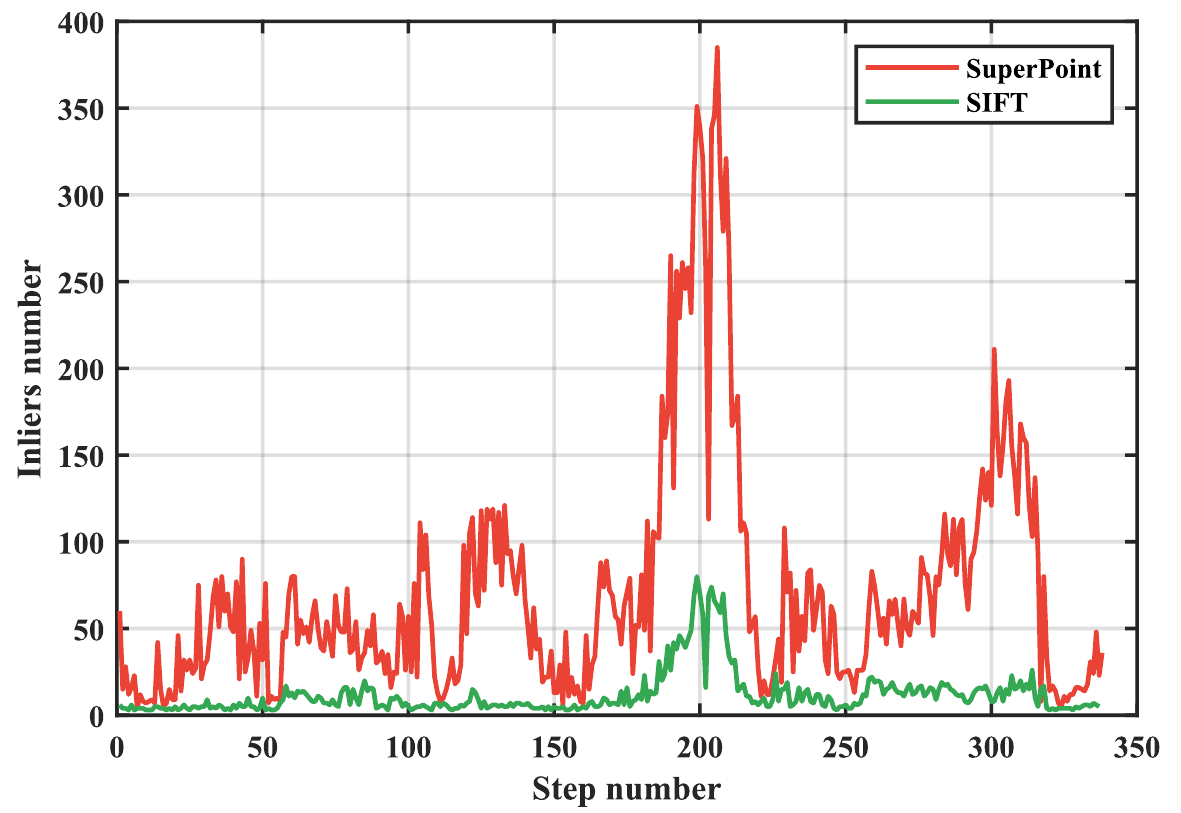}
  \caption{The number of inliers comparison between SuperPoint feature and SIFT features in nighttime environment.}
  \label{fig11}
\end{figure}

\subsection{Outdoor Experiment in Low-Light Condition}
To further assess the robustness of our method in visually challenging environments, we conducted a third experiment under outdoor nighttime conditions. Figure~\ref{fig10} showcases an example of the nighttime images captured using a mobile phone. The low lighting conditions during nighttime result in poor image quality, presenting a significant challenge for visual tracking methods.
Traditional handcrafted feature descriptors exhibit limited robustness in such challenging environments due to their poor invariance. In this study, we introduced learned features~\cite{SuperPoint} into our image-based pipeline. These learned features generate denser and more accurate matches compared to traditional methods like SIFT, as depicted in Figure~\ref{fig11}. This integration effectively enhances the reliability and continuity of conventional visual methods.

\begin{figure}[t]
  \centering
  \includegraphics[width=0.8\hsize]{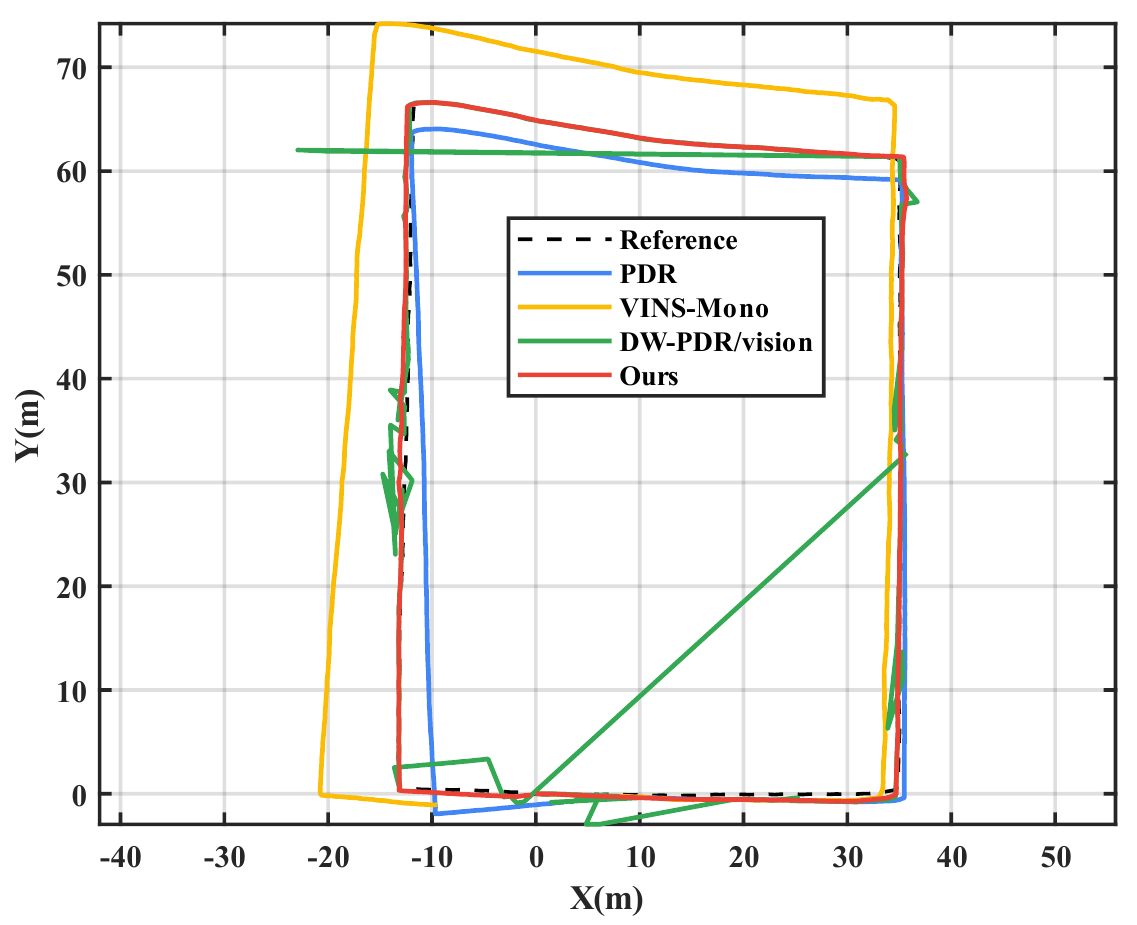}
  \caption{The trajectory comparison of different methods in outdoor nighttime environment.}
  \label{fig12}
\end{figure}

\begin{figure}[t]
  \centering
  \includegraphics[width=0.8\hsize]{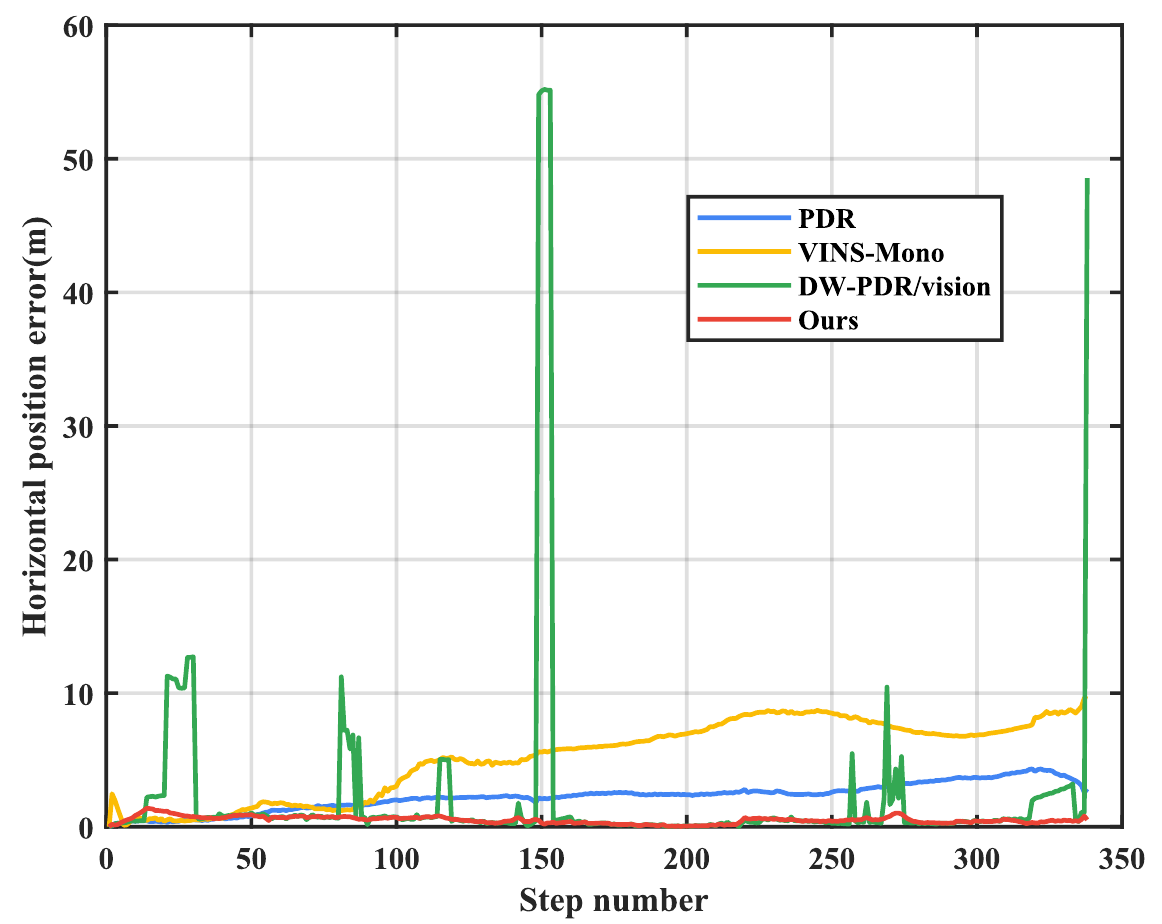}
  \caption{Distribution of horizontal position error comparison with the change of detected steps.}
  \label{fig13}
\end{figure}

\begin{figure}[t]
  \centering
  \includegraphics[width=0.8\hsize]{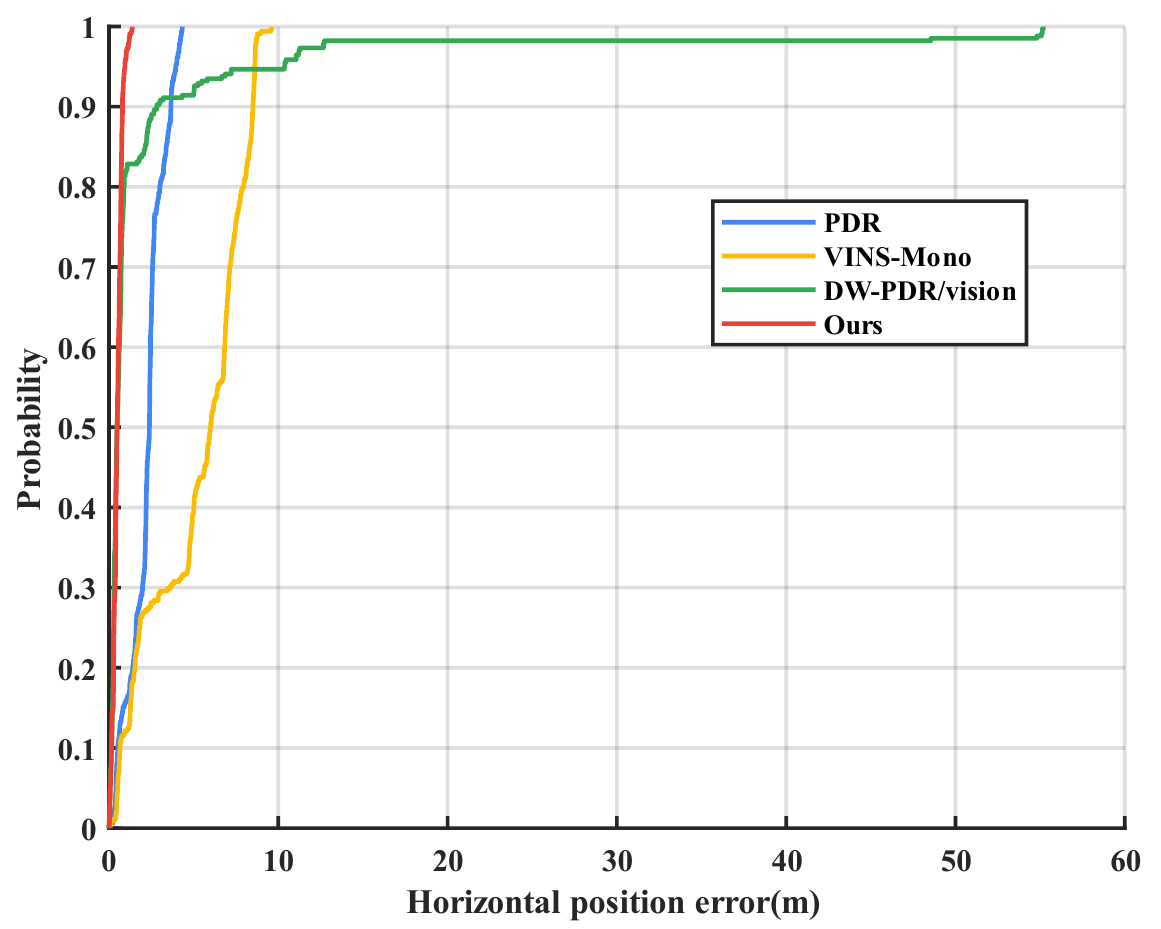}
  \caption{Cumulative distribution of horizontal positioning errors across different methods.}
  \label{fig14}
\end{figure}

Figure~\ref{fig12} displays the trajectory results of different approaches in an outdoor nighttime environment. It is evident that our method consistently provides continuous and accurate trajectory estimations, showcasing robustness against severely degraded environments.
The pure inertial-based PDR approach yields acceptable results as inertial sensors are independent of visual cues. However, its positioning error accumulates over time.
In the nighttime environment, VINS-Mono's positioning results deviate significantly from the reference trajectory due to the lack of sufficient features for accurate match tracking.
Although the incorporation of learned features improves the robustness of positioning for nighttime query images to some extent, visual relocalization still encounters numerous failures due to the limited representation of local features.
Figure~\ref{fig13} illustrates the horizontal positioning errors with significant fluctuations observed in the DW-PDR/vision method. This is attributed to relying solely on the number of inliers as a criterion for evaluating reliability, which can introduce abnormal visual results. The dynamic weighting fusion strategy fails to eliminate the impact of incorrect observations, resulting in estimated trajectories deviating significantly from the reference path.
In contrast, our method effectively addresses the risk of visual relocalization failures and mitigates the impact of abnormal observations on positioning accuracy by employing an optimization-based fusion strategy. Additionally, our method achieves globally smooth and drift-free trajectory estimations through incremental smoothing optimization.

Figure~\ref{fig14} displays the cumulative distribution of horizontal positioning errors, demonstrating the superior performance of our method compared to other algorithms. Table~\ref{tab3} provides statistical analysis results of horizontal positioning errors in outdoor nighttime conditions. Our method improves the positioning accuracy of PDR by 77.0\% and reduces the maximum error from 4.3309 m to 1.3854 m. Remarkably, even in the dark lighting environment, our method outperforms VINS-Mono in terms of positioning accuracy, which is surprising given the latter's reliance on visual information.
The positioning accuracy of DW-PDR/vision is significantly affected by abnormal visual observations. Conversely, our method consistently achieves accurate and remarkable positioning results, with a loop error of only 0.5152 m, underscoring its robustness against visually challenging environments.

\begin{table}[t]
  \renewcommand\arraystretch{1.25}
  \centering
  \caption{Comparison of positioning accuracy in outdoor nighttime experiment}
  \label{tab3}
  \begin{tabular}{p{55pt}<{\centering}p{48pt}<{\centering}p{48pt}<{\centering}p{48pt}<{\centering}}
    \hline
     & RMSE(m) & Max Error(m) & Loop Error(m) \\ \hline
    PDR & 2.4589 & 4.3309 & 2.8320 \\ 
    VINS-Mono & 5.9169 & 9.6373 & 9.5584 \\ 
    DW-PDR/vision & 7.6334 & 55.1899 & 48.5443 \\ 
    Ours & \textbf{0.5655} & \textbf{1.3854} & \textbf{0.5152} \\ 
    \hline
  \end{tabular}
\end{table}

\section{Conclusion}
To achieve self-reliant and robust pedestrian navigation using a smartphone in visually challenging environments, our work proposes, ReLoc-PDR, a robust pedestrian positioning framework that integrates Pedestrian Dead Reckoning (PDR) and visual relocalization based on incremental smoothing optimization.
Considering the visual degradation problem in environments with weak textures and varying illumination, we introduce a visual relocalization pipeline using the learned features from deep neural network instead of traditional handcrafted features. This effectively establishes 2D-3D correspondences with higher inlier rates, enhancing the robustness of the pedestrian localization system.
Furthermore, we propose an optimization-based fusion strategy that couples the PDR and visual relocalization poses into a graph model. This fusion strategy is accompanied by the use of the Tukey robust kernel, which helps eliminate the risk of abnormal visual observations.
Experimental results demonstrate the effectiveness of our ReLoc-PDR in various challenging environments, including corridors with limited texture, overcast weather conditions, and dark nighttime scenarios. The proposed ReLoc-PDR method achieves accurate and smooth trajectory estimation, continuously providing pedestrian positions at a decimeter-level accuracy and a high frequency. 

\bibliographystyle{unsrt}
\bibliography{ref}

\begin{IEEEbiography}[{\includegraphics[width=1in,height=1.25in,clip,keepaspectratio]{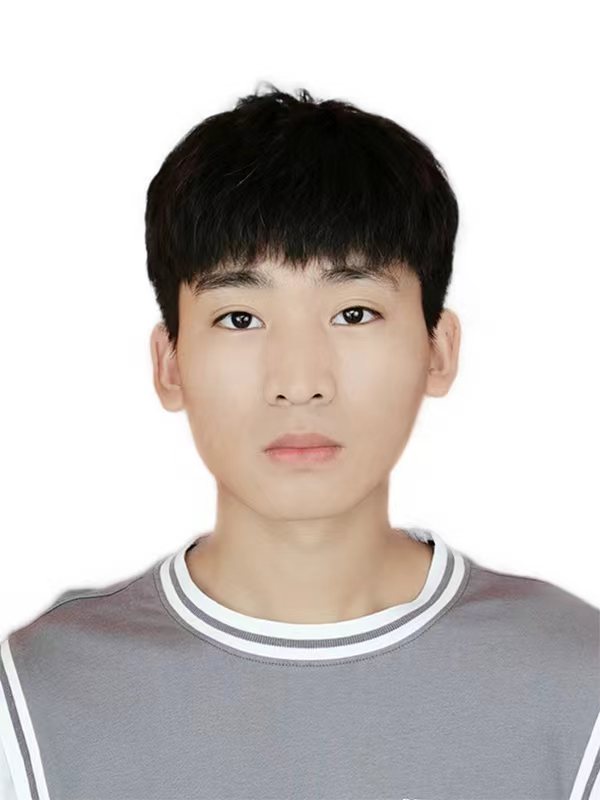}}]{Zongyang Chen} received the B.S. degree from Fuzhou University, Fuzhou, China, in 2021. He is currently pursuing the master’s degree with the National University of Defense Technology, Changsha, China. His research interests include pedestrian inertial navigation and visual–inertial integrated navigation.
\end{IEEEbiography}

\begin{IEEEbiography}[{\includegraphics[width=1in,height=1.25in,clip,keepaspectratio]{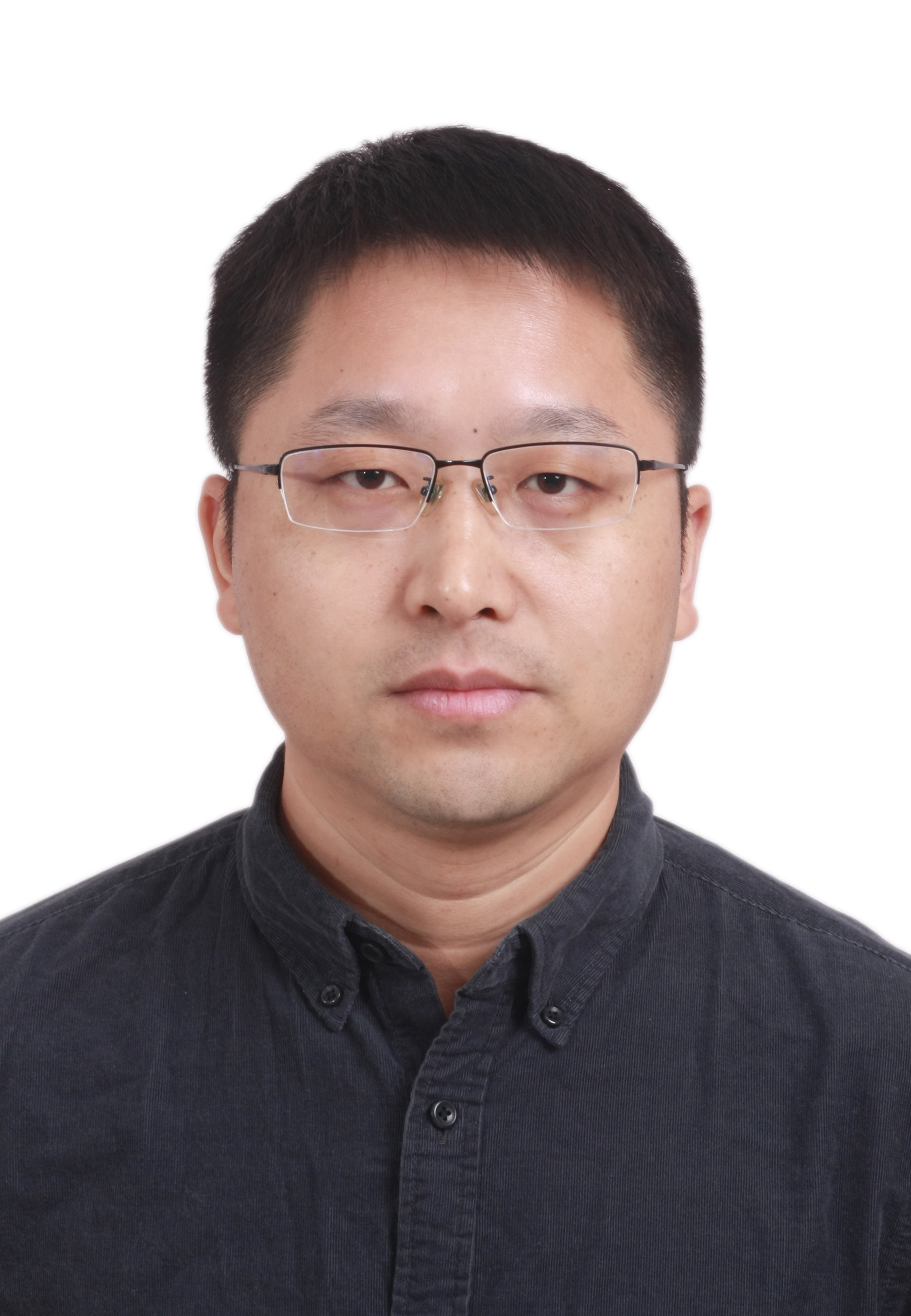}}]{Xianfei Pan} received the Ph.D. degree in control science and engineering from the National University of Defense Technology, Changsha, China, in 2008. Currently, he is a professor of the College of Intelligence Science and Technology, National University of Defense Technology. His current research interests include Inertial navigation system and indoor navigation system.
\end{IEEEbiography}

\begin{IEEEbiography}
[{\includegraphics[width=1in,height=1.25in,clip,keepaspectratio]{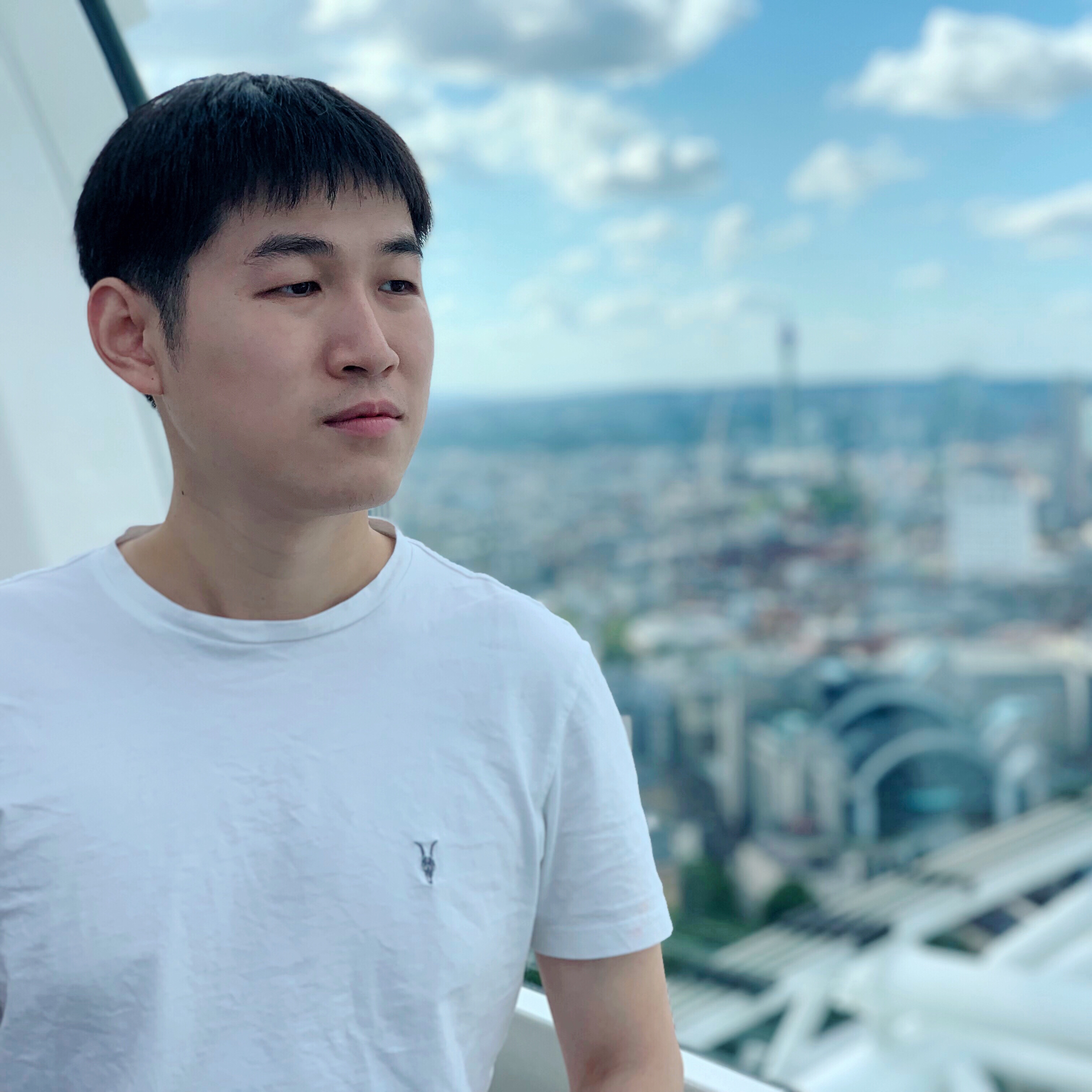}}]{Changhao Chen} obtained his Ph.D. degree at University of Oxford (UK), M.Eng. degree at National University of Defense Technology (China), and B.Eng. degree at Tongji University (China). Now he is a Lecturer at College of Intelligence Science and Technology, National University of Defense Technology (China). His research interest lies in robotics, computer vision and cyberphysical systems. 
\end{IEEEbiography}

\end{document}